\pdfoutput=1

\documentclass[11pt]{article}

\usepackage[]{acl}

\usepackage{times}
\usepackage{latexsym}
\usepackage{float}
\restylefloat{table}
\usepackage[T1]{fontenc}

\usepackage[utf8]{inputenc}

\usepackage{microtype}
\usepackage{comment}

\usepackage{multirow}
\usepackage{multicol}
\usepackage{graphicx}
\usepackage{physics}
\usepackage{booktabs}
\makeatletter

\usepackage{amssymb}

\newcommand{\STAB}[1]{\begin{tabular}{@{}c@{}}#1\end{tabular}}

\title{On Uncertainty Calibration and Selective Generation in Probabilistic Neural Summarization: A Benchmark Study}

\author{ Polina Zablotskaia\thanks{~~Equal contribution.} \quad Du Phan  \footnotemark[1] \quad Joshua Maynez\quad Shashi Narayan \\ {\bf Jie Ren } \quad {\bf Jeremiah Liu}\\
Google Research\\
\texttt{\small{\{polinaz,phandu,joshuahm,shashinarayan,jjren,jereliu\}@google.com}}
}

\begin{document}
\maketitle
\begin{abstract}

Modern deep models for summarization attains impressive benchmark performance, but they are prone to generating \textit{miscalibrated} predictive uncertainty. This means that they assign high confidence to low-quality predictions, leading to compromised reliability and trustworthiness in real-world applications. Probabilistic deep learning methods are common solutions to the miscalibration problem. However, their relative effectiveness in complex autoregressive summarization tasks are not well-understood. In this work, we thoroughly investigate different state-of-the-art probabilistic methods' effectiveness in improving the uncertainty quality of the neural summarization models, across three large-scale benchmarks with varying difficulty.
We show that the probabilistic methods consistently improve the model's generation and uncertainty quality, leading to improved selective generation performance (i.e., abstaining from low-quality summaries) in practice. We also reveal notable failure patterns of  probabilistic methods widely-adopted in NLP community (e.g., Deep Ensemble and Monte Carlo Dropout), cautioning the importance of choosing appropriate method for the data setting.
\end{abstract}

\section{Introduction}
\vspace{-0.3em}
In recent years, autoregressive deep models for text summarization have achieved impressive performance. However, despite their success, these models often suffer from a critical flaw: they generate prediction with high confidence even when the quality of the summary is low \citep{xu2022sequence}. This can severely compromise the reliability and trustworthiness of the generated summaries in real-world applications. In the probabilistic forecastt literature, such issue is known under the term \textit{miscalibration}, i.e., the model's predictive confidence is mis-aligned with its prediction quality. For example, in classification tasks, a model is said to be \textit{miscalibrated} if for all test examples where it predicts with probability 0.9, the model's actual accuracy for these examples deviates far from 90\% \citep{guo2017calibration, gneiting2007probabilistic}. Despite its practical importance, this notion of \textit{uncertainty calibration} has received much less attention in the summarization literature until recently, with the proposed techniques mostly focusing on training deterministic models
\citep{cao2021cliff,xu2022sequence,sun2021alleviating, zhao2022calibrating, liu2022brio, jung2021moment}.

In the uncertainty literature, probabilistic deep learning has emerged as a principled approach to tackle model miscalibration while maintaining prediction quality \citep{nado2021uncertainty}. Intuitively, probabilistic DNNs generates and multiple plausible predictions from its posterior predictive $\bar{p}_m(y|x) = \frac{1}{M} \sum_{m=1}^M p_m(y|x)$ and report the average, thereby mitigating the overconfidence of the individual model prediction.
Although well-tested in classification tasks, the effectiveness of different state-of-art probabilistic methods in improving neural summarization models' uncertainty quality has been less explored. The existing study mostly focuses on a particular classic method (e.g., Monte Carlo Dropout, MCD) and tested on relatively simple datasets that doesn't fully capture the realistic usage \citep{gidiotis2022should}.

In this work, we address this by conducting a comprehensive investigation of the relative effectiveness of state-of-the-art probabilistic methods in improving the uncertainty quality of neural summarization models. We interrogate both classic approaches such as Monte Carlo Dropout (MCD) and Deep Ensemble (DE), and more recent state-of-art methods such as Batch Ensemble (BE) Spectral-normalized Gaussian process (SNGP) and their combinations that address the latecy and quality caveats of the classic methods \citep{gal2016dropout, lakshminarayanan2017simple, liu2020simple, wen2020batchensemble}. 
Furthermore, we evaluate method performance across multiple benchmarks of varying difficulty to ensure the practical relevance of our result, and to uncover potential failure patterns of different approaches. Our contributions are:\\
$\bullet$ We adapt the various probabilistic deep learning methods to the LLM setup and conduct an extensive study on their effect on both uncertainty and prediction aspects of model performance. \\
$\bullet$ We propose evaluation metrics to measure the uncertainty calibration performance of summarization models, tailored toward domain-specific quality scores (e.g., ROGUE).\\
$\bullet$ We show that using probabilistic methods generally leads to improved summarization and calibration performance, and consequently improved selective generation. We also discuss the failure patterns of the popular methods such Deep Ensembles \citep{lakshminarayanan2017simple} and Monte Carlo Dropout \citep{gal2016dropout}. 

\section{Related work}
\vspace{-0.3em}
\paragraph{Probabilitistic learning for seq2seq models} 
Developed primarily in the context of discriminative models,
the state-of-art probabilistic approaches can be applied to large neural models without sacrificing performance \citep{gal2016dropout, lakshminarayanan2017simple, wen2020batchensemble, liu2020simple}.
Recently, however, initial investigations into unsupervised uncertainty estimation for structured prediction have appeared, primarily focusing on more basic approaches such as Monte Carlo Dropout (MCD) or Deep Ensemble (DE)
\citep{xiao2019quantifying,wang2019improving, fomicheva2020unsupervised, Malinin2021UncertaintyEI, lin2022neural}, with a few work looking into summarization tasks \citep{xu2020understanding, gidiotis2022should}. In comparison, this work focuses on an unbiased evaluation of a wide range of state-of-the-art probabilistic methods on tasks with varying difficulty, and reveals failure patterns of classic approaches such as MCD and DE. 
\paragraph{Calibration Technique in Language Processing}
\citet{guo2017calibration} proposed improving calibration of document classifier using of temperature scaling. \citet{muller2019does} and \citet{wang-etal-2020-inference} explored improving calibration in neural machine translation using label smoothing. \citet{desai-durrett-2020-calibration} noted that calibration methods can be used to improve the accuracy of pre-trained language models. 
\citet{jung2021moment} proposed a novel training approach to improve calibration by minimizing a combined loss of cross-entropy and calibration
In the summarization literature,  \citep{cao2021cliff,xu2022sequence,sun2021alleviating, zhao2022calibrating, liu2022brio} explored calibrating model probability using contrastive learning approaches. Most of these techniques focus on deterministic models. They are orthogonal to and can be combined with the probabilistic approaches we explore in this work.

\vspace{-0.3em}
\section{Methods}
\label{sec:method}
\vspace{-0.3em}
Probabilistic methods have been adopted to increase the reliability of large language models. Plex paper \citep{tran2022plex} provided a nice survey on the robustness of uncertainty methods on text classification tasks. We employ the following techniques to assess their impact on improving calibration in summarization.\vspace{0.05in}\\
\textbf{Single-model methods:}\vspace{0.05in}\\
$\bullet$ \textbf{Deterministic Baseline} - we use the base T5 model \footnote{All methods can be applied to larger models.} \citep{raffel2020exploring} as the baseline model.\\
$\bullet$ \textbf{Monte Carlo Dropout (MCD)} \citep{gal2016dropout} which estimates uncertainty using the Monte Carlo average of 10 dropout samples. Those samples are generated using the same model and parameters but with different random seeds at dropout layers.\\
$\bullet$ \textbf{Batch Ensemble (BE)} \citep{wen2020batchensemble} - an ensemble method which has much lower computational costs comparing to MC Dropout and Deep Ensemble. We replace the last transformer's MLP block by a batch ensemble block with ensemble size be 5\footnote{BE requires more memory on a single machine, so we keep the ensemble size below 10.}.\\
$\bullet$ \textbf{Spectral-normalized Neural Gaussian Process (SNGP)} \citep{liu2020simple} - a recent state-of-the-art approach which improves uncertainty quality by transforming a neural network into an approximate Gaussian process model. The Gaussian Process last layer is able to reflect the distance between a test example and the training set, hence potentially be helpful in improving calibration.\\
$\bullet$ \textbf{SNGP+MCD} which is the MC Dropout on top of an SNGP model;
\vspace{0.05in}\\
\textbf{Multi-model methods:}\vspace{0.05in}\\
$\bullet$ \textbf{Deep Ensemble (DE)} \citep{lakshminarayanan2017simple} which trains 10 deterministic models individually and averages all. We use the same model architecture but changing the initial seeds.\\
$\bullet$ \textbf{Gaussian Process Ensemble (SNGP+DE)} is the combination of deep ensemble and SNGP. 

For all methods, we use the official base T5 checkpoint, which are pretrained on a large corpus like C4 \citep{raffel2020exploring}. We then finetune the parameters on summarization tasks.
To generate prediction from the model posterior, we perform beam inference with respect to the model's conditional posterior mean probability, i.e., $\bar{p}(y_t|y_{<t},x)=\frac{1}{M}\sum_{m=1}^M p_m(y_t|y_{<t},x)$, where $M=10$ is the number of samples from model posterior.
To quantify model uncertainty, we consider the length-normalized predicted log-probabilities following previous work, i.e., $u(y|x) := \frac{1}{T} \sum_{t=1}^T \bar{p}(y_t|y_{<t},x)$ \citep{wu2016google, liu2022brio}.

\begin{figure*}
    \centering
    \includegraphics[width=0.85\textwidth]{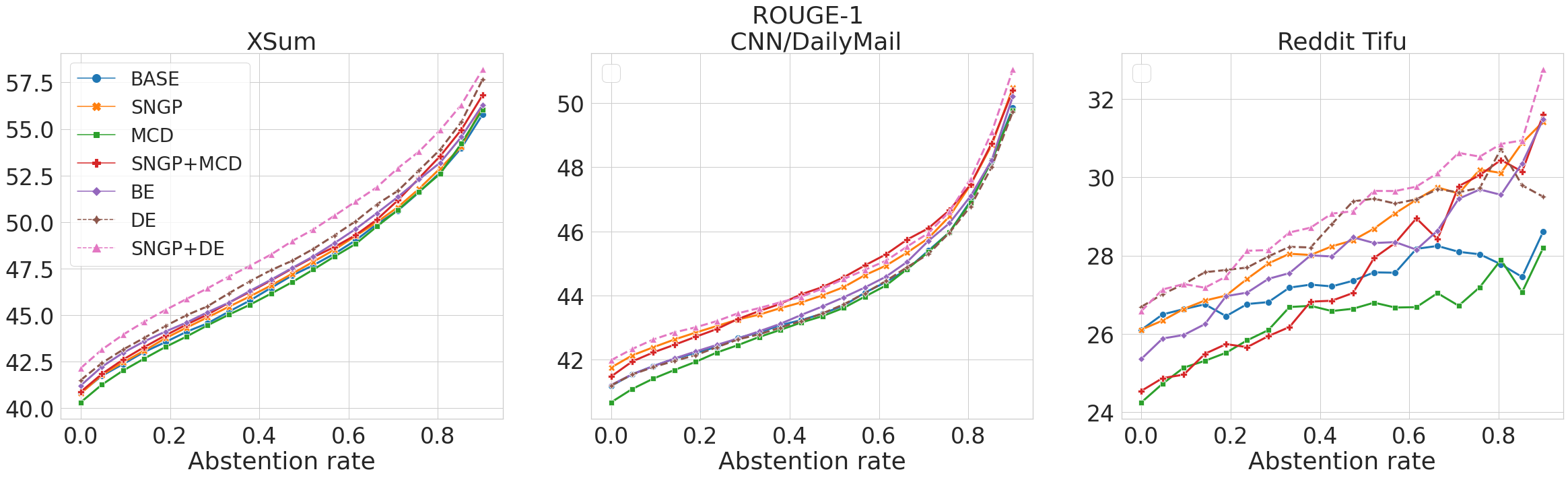}
    \caption{Quality vs Abstention Curve for different probablistic methods. 
    \textbf{Abstention rate $\alpha$} denotes the percentage of examples that were excluded, after ranking according to log-probabilities. 
    For single model methods (solid lines), SNGP+MCD models have generally higher ROUGE scores in CNN/DM, and in regions of $\alpha>0.6$ in XSUM and Reddit.  
    For multi-model methods, SNGP+DE generally outperforms DE in all the three datasets. 
    }
    \label{fig:abst_main}
    \vspace{-1em}
\end{figure*}

\section{Experiments}
\vspace{-0.3em}
\subsection{Datasets}
\textbf{XSUM} \citep{narayan2018don} consists of 227k BBC articles from 2010 to 2017 with a single sentence highly abstractive summary. Sometimes the summary contains information not present in the article. 
\\
\textbf{CNN/DailyMail} \citep{hermann2015teaching, see2017get} contains 313k articles from the CNN and Daily Mail newspapers with bullet point summaries. The summaries are on  average 3-4 sentences and relatively extractive. 
\\
\textbf{RedditTIFU-long} \citep{kim-etal-2019-abstractive} contains 42k posts of informal stories from sub-reddit TIFU from 2013-Jan to 2018-Mar with author written summaries. The style and length of the summaries are very diverse. 
\vspace{-0.3em}
\subsection{Summarization quality using probabilistic methods}
\vspace{-0.3em}
We first study how well different probabilistic methods on summary prediction, comparing with the baseline deterministic model. 
We use ROUGE-1/2/L \citep{lin-2004-rouge}
to measure general summarizaiton quality. As shown in Table \ref{tab:rouge}, we observe the consistent improvement of the the ROUGE scores in probabilistic models compared to baselines. 
For single model methods, SNGP achieves the highest average ROUGE scores over the three datasets. 
Other probabilistic methods also show promising performance: SNGP+MCD is ranked the second top regarding ROUGE-1, and BE is ranked the second top regarding ROUGE-2 and the top regarding ROUGE-L. 
For multiple model methods, SNGP+DE improves over the deterministic DE. 
Comparing multiple model methods with single model methods, DE and SNGP+DE generally have higher ROUGE scores than single model methods. 

\begin{table}[ht]
\captionsetup{font=footnotesize}
\centering
\vspace{-0.5em}
\resizebox{0.39\textwidth}{!}{%
\begin{tabular}{cccc|cc}
\toprule
\multicolumn{6}{c}{\textbf{ROUGE-1}$\uparrow$}\\
Method & XSUM & CNN/DM & Reddit & Average$\uparrow$ & Average Rank$\downarrow$ \\ 
\hline
Base     & 40.83   & 41.19 & 26.14 & \underline{36.05}     & 3.00 \\
SNGP     & 40.79   & 41.76 & 26.08 & \textbf{36.21}  & \textbf{2.00} \\
MCD      & 40.31   & 40.68 & 24.27 & 35.09 & 5.00 \\
SNGP+MCD & 40.90    & 41.49 & 24.60  & 35.66    & \underline{2.33} \\
BE       & 41.21   & 41.22 & 23.36 & 35.26    & 2.67 \\ \midrule
DE       & 41.51   & 41.20  & 26.65 & \underline{36.45}    & \underline{1.67}  \\
SNGP+DE  & 42.14   & 41.99 & 26.57 & \textbf{36.90}      & \textbf{1.33}      \\  
\toprule\multicolumn{6}{c}{\textbf{ROUGE-2}$\uparrow$}\\ 
Method & XSUM & CNN/DM & Reddit & Average$\uparrow$ & Average Rank$\downarrow$ \\ 
\hline
Base     & 19.14   & 19.77 & 7.76  & \underline{15.56}  & 2.67   \\
SNGP     & 18.91   & 20.19 & 7.76  & \textbf{15.62} & \textbf{2.00}  \\
MCD      & 18.63   & 19.78 & 7.12  & 15.18 & 4.00 \\
SNGP+MCD & 18.91   & 20.33 & 7.00     & 15.41 & 3.00    \\
BE       & 19.41   & 19.81 & 7.31  & 15.51 & \underline{2.33} \\ \midrule
DE       & 19.84   & 19.77 & 8.21 & \textbf{19.81} & \underline{1.67}    \\
SNGP+DE  & 20.35   & 20.49 & 7.77  & \underline{16.20} & \textbf{1.33}  \\
\toprule\multicolumn{6}{c}{\textbf{ROUGE-L}$\uparrow$}\\ 
Method & XSUM & CNN/DM & Reddit & Average$\uparrow$ & Average Rank$\downarrow$ \\ 
\hline
Base     & 33.76   & 38.54 & 21.31 & \underline{31.20} & \underline{2.33} \\
SNGP     & 33.53   & 39.12 & 21.17 & \textbf{31.27} & 2.67 \\
MCD      & 33.21   & 38.09 & 20.06 & 30.45 & 5.00  \\
SNGP+MCD & 33.59   & 38.97 & 20.00    & 30.85 & 3.00 \\
BE       & 34.06   & 38.50  & 20.81 & 31.12  & \textbf{2.00} \\ \midrule
DE       & 34.38   & 38.54 & 21.72 & \textbf{31.55}  &  \textbf{1.33} \\
SNGP+DE  & 34.92   & 36.55 & 21.18  & \underline{30.88} & 1.67 \\ 
\bottomrule
\end{tabular}
}
\caption{
ROUGE scores and ranking of different probabilistic methods across all datasets. Probablistic methods consistently outperform base model, and SNGP-family models generally lead to strong performance.}
\label{tab:rouge}
\end{table}

\vspace{-1.5em}
\subsection{Measuring Uncertainty Calibration in Summarization}
We now study model's uncertainty calibration quality. We consider both the classic metric Expected Calibration Error (ECE), and also 
the uncertainty score's Spearman's rank correlation with  domain-specific quality scores tailored for summarization (e.g., ROGUE). 
\\
\textbf{Expected Calibration Error (ECE).}
In order to evaluate whether the model estimated probabilities have been more calibrated we access the difference in expectation between confidence and accuracy using ECE metric \citep{naeini2015obtaining}:
\vspace{-1.5em} 
\begin{center}
\small
$$ECE = \sum_{k=1}^{K}\frac{|B_k|}{n}|\mathrm{conf}(B_k) - \mathrm{acc}(B_k)|,$$
\end{center}
\vspace{-0.5em}

where we split the interval $(0, 1]$ into $K$ equal-size bins and define $B_k$ to be the set containing the indices of examples which have predicted probability lie in the $k$-th bin: $B_k =\left\{ i |\hat{p}_i \in \left(\frac{k-1}{K}, \frac{k}{K}\right]\right\}$, where the average accuracy and confidence within each bin are defined as $\mathrm{acc}(B_k) = \frac{1}{|B_k|} \sum_{i\in B_k} I(\hat{y}_i=y_i)$ and $\mathrm{conf}(B_k) = \frac{1}{|B_k|}\sum_{i\in B_k} \hat{p}_i$. 
In auto-regressive prediction, $\hat{y}_i$ can be a sequence or a token\footnote{During evaluation, we compute token probabilities from the highest scoring beam sequence.}, which corresponds to two different metrics \emph{sequence-level ECE} and \emph{token-level ECE} respectively. 
As shown in Table \ref{tab:ece}, across all methods, SNGP+MCD and SNGP+DE generally leads to lower ECE in single model and multi-model methods respectively, suggesting SNGP helps to reduce ECE. 


\begin{table}[ht]
\captionsetup{font=footnotesize}
    \centering
    \resizebox{0.5\textwidth}{!}{%
    \begin{tabular}{ccccc|cccc}
    \hline
  & \multicolumn{4}{c}{\textbf{Sequence-level ECE$\times 10^{-3}$ $(\downarrow)$ } } &  \multicolumn{4}{c}{ \textbf{Token-level ECE$\times 10^{-1}$ $(\downarrow)$ }}  
  \\ 
  \textbf{Method}& XSUM & CNN/DM & Reddit & Average & XSUM & CNN/DM & Reddit & Average\\ \hline
 
 Base & 2.70& 0.28 & 0.32 & 1.10 & 5.89  & 7.69  & 4.56 & 6.05\\ 
 SNGP  & 3.47 & 0.52 & 1.13 & 1.71 & 5.97 & 7.71 & 5.26 & 6.31 \\ 
 MCD & 1.02  & 0.18 & 0.05 & \underline{0.42} & 4.54 & 6.68 & 3.05 & \textbf{4.76} \\ 
 SNGP+MCD & 0.93 & 0.11  & 0.13 & \textbf{0.39} & 4.54 & 6.69 & 3.44 & \underline{4.89}\\ 
 BE & 2.65 & 0.64 &0.44 & 1.24 & 5.95 & 8.04 & 5.01  & 6.33 \\ \hline
      DE & 1.89 & 0.13 & 0.63 & \underline{0.88} & 5.58 & 7.78  & 4.82 & \underline{6.06} \\ 
 SNGP+DE & 0.83 & 0.36 & 0.90 & \textbf{0.70} & 5.39 & 7.62 & 5.15 & \textbf{6.05} \\
  \hline
    \end{tabular}
    }
    \caption{ECE on sequence and token levels of different probabilistic methods across all datasets. SNGP+MCD and SNGP+DE generally leads to lower ECE in single model and multi-model methods, respectively.}
    \label{tab:ece}
\end{table}

\vspace{-1em}
 
\begin{figure}[ht]
\captionsetup{font=footnotesize}
\centering
\resizebox{0.5\textwidth}{!}{%
\includegraphics[width=0.5\textwidth]{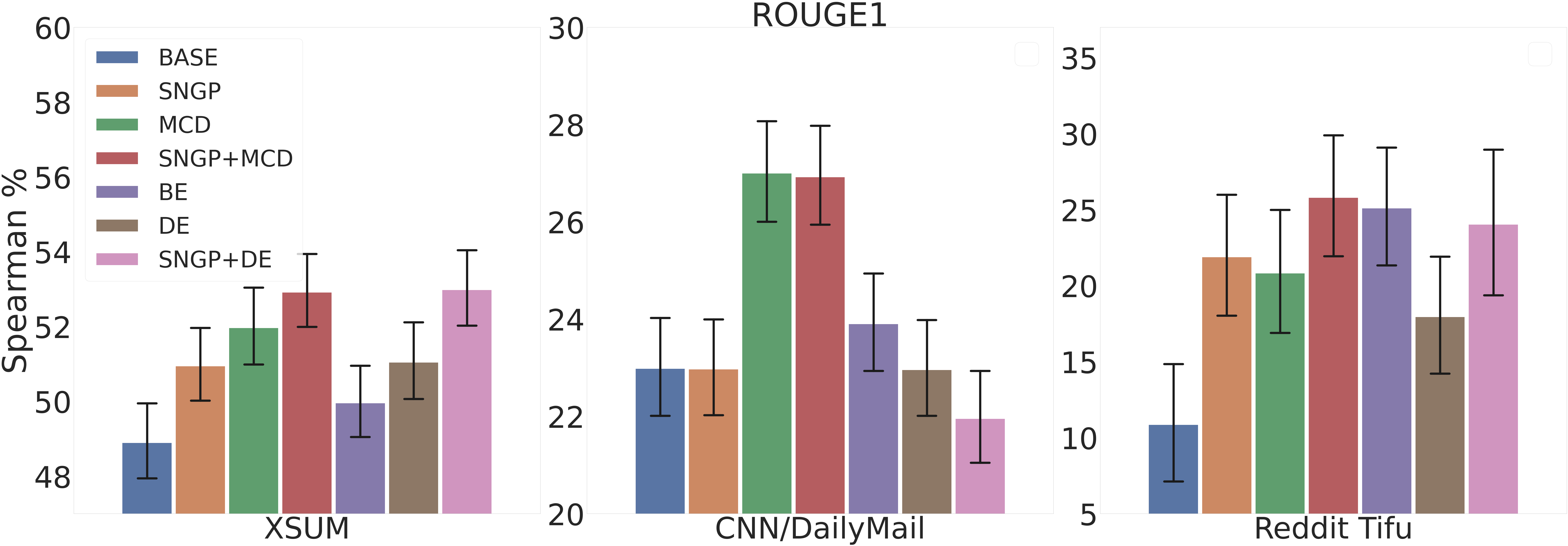}
}
\caption{Spearman's rank correlation between the length-normalized log-probabilities and the ROUGE-1. We compute the error bars using bootstrap standart deviation technique.}
\label{fig:spearman}
\vspace{-1.75em}
\end{figure}

\paragraph{Rank Correlation with quality score}
We investigate how the Spearman's rank correlation between the log-probabilities and ROUGE changes with calibration. 
Overall we see a general trend demonstrating the calibration increases the correlation, as shown in Figure \ref{fig:spearman}. For the ROC-AUC scores please refer to the section \ref{sec:roc}.

\subsection{Selective Generation via Abstention}
Selective generation refers to the procedure to
selectively generate higher-quality outputs while abstain the low-quality outputs \cite{ren2022out}. It evaluates the models' uncertainty calibration, because a well calibrated model is expected have high uncertainty for low-quality outputs such it can be used to select examples to abstain.
We use the \textit{Quality vs Abstention Curve} to compare methods: specifically, at a given abstention rate $\alpha$, we remove the lowest $\alpha$-fraction uncertain examples and compute the average quality of the remaining examples, as shown in Figure \ref{fig:abst_main}.
For single model methods (solid lines), SNGP+MCD models have generally higher ROUGE scores in CNN/DM, and in regions of $\alpha>0.6$ in XSUM and Reddit.  
For multi-model methods, SNGP+DE generally outperforms DE in all the three datasets. \\
\textbf{Failure Patterns.}
When comparing multi-model methods with single model methods, 
we observe that XSUM and Reddit both have multi-model methods outperforming single model methods, but CNN/DM does not benefit from using multi-model methods.
This difference can be explained by the fact that CNN/DM is an simpler task that is more extractive in nature, and a single model already performs well and  relatively calibrated. In this case, using a deep ensemble can in fact lead to under-confidence \citep{rahaman2021uncertainty}. Furthermore, in Reddit dataset, MCD-family methods seem to lead to severe degradation of summarization quality. Note that Reddit is a more challenging task with much greater linguistic diversity when compared to XSUM and CNN/DailyMail, cautioning the use of MCD method in challenging test environments where a single model does not perform well. 

\section{Conclusion} 
\vspace{-0.3em}
We conduct an extensive study of the popular probabilistic deep learning calibration methods  applied to LLM. Consistently positive effect of these techniques is reflected in improved performance in summarization quality, uncertainty calibration, and selective generation. 
\newpage
\section{Limitations} 
In our paper we investigated the effect of most common and widely used probabilistic deep learning methods. Even though we observed a positive effect of calibration on the variety of metrics, that impact wasn't strong enough to change the course of LLM reliability on a large scale and we don't know whether our findings would be  general across larger models and different tasks. One explanation as to why the we didn't observe higher level of calibration could be ground in the LLMs training objective. Maximum Likelihood Estimation (MLE)  tends to overfit to the training data, because  models are faced with a single ground-truth example per input text. In other words, LLMs trained with MLE are more likely to assign high probabilities to examples from the training data, even when they are not representative of the true distribution of the language. Further research is needed to understand how we can best adapt the learning objective in order to take the most advantage from the probabilistic deep learning methods.
\section{Ethical impact}
Our work directly contributes to the topic of reliable deep learning. We believe our work should positively affect the scientific community, since we address one of the main problems that often occurs in the machine learning research: how to make the models more reliable and trustworthy. We hope that in long run we can arrive at a standardized benchmark set of techniques that can help the NLP community develop LLMs that are universally trustworthy.


\bibliography{anthology,custom}
\bibliographystyle{acl_natbib}

\clearpage
\onecolumn
\appendix
\section{Appendix}
\subsection{ROC-AUC scores}
\label{sec:roc}
\begin{table}[!ht]
    \centering
    \resizebox{0.48\textwidth}{!}{%
    \begin{tabular}{cccccc}
 \textbf{Metric} & \textbf{Method Type} & \textbf{Method} & \textbf{XSUM} & \textbf{CNN/DailyMail} & \textbf{Reddit Tifu} \\ \hline
       \multirow{7}*{\STAB{\rotatebox[origin=c]{90}{ROUGE1}}} &\multirow{5}*{Single-model}&Base & 73.16 & 60.38 & 60.28 \\ 
 &&SNGP & 74.34 & 59.96 & 65.01 \\ 
 &&MCD & 75.06 & 62.29 & 69.98 \\ 
 &&SNGP+MCD & 75.14 & 62.05 & 67.60 \\ 
 &&BE & 73.30 & 60.45 & 64.08 \\\cline{2-6}
 &\multirow{2}*{Multi-model (10)}&DE & 73.90 & 59.97 & 64.07 \\ 
 &&SNGP+DE & 75.19 & 59.25 & 63.65 \\ \hline
 \multirow{7}*{\STAB{\rotatebox[origin=c]{90}{ROUGE-2}}}&\multirow{5}*{Single-model}&Base & 73.00 & 60.22 & 59.93 \\ 
 &&SNGP & 73.46 & 60.47 & 63.96 \\ 
 &&MCD & 74.57 & 62.38 & 61.27 \\ 
 &&SNGP+MCD & 74.58 & 63.22 & 63.58 \\ 
 &&BE & 72.82 & 59.90 & 63.23 \\  \cline{2-6}
 &\multirow{2}*{Multi-model (10)}&DE & 74.03 & 59.93 & 62.59 \\ 
 &&SNGP+DE & 74.52 & 60.27 & 61.79 \\  \hline
 \multirow{7}*{\STAB{\rotatebox[origin=c]{90}{ROUGE-L}}}&\multirow{5}*{Single-model}&Base & 71.75 & 59.25 & 61.34 \\ 
 &&SNGP & 72.86 & 59.06 & 59.44 \\ 
 &&MCD & 73.86 & 61.38 & 66.33 \\ 
 &&SNGP+MCD & 74.61 & 61.35 & 62.99 \\ 
 &&BE & 72.45 & 60.03 & 64.42 \\  \cline{2-6}
 &\multirow{2}*{Multi-model (10)}&DE & 72.85 & 58.78 & 65.05 \\
 &&SNGP+DE & 74.15 & 59.51 & 61.39 \\ \hline
    \end{tabular}}
    \caption{We measure the Area Under Curve values, when using the log-probabilities as a signal for "good/bad" summaries. Good and bad summaries are defined by a threshold $\theta$ we impose on the metric, i.e.  when metric is above certain $\theta$ then we treat the summary as good and when it is below we treat it as a bad summary. We used the following $\theta$ for ROUGE1, ROUGE-2 and ROUGE-L correspondingly: 40, 15 and 30.}
    \label{tab:roc_auc}
\end{table}

\subsection{Spearman's rank correlation ROUGE-2 and ROUGE-L}
\begin{figure*}[ht]
    \centering
    \resizebox{0.9\textwidth}{!}{%
    \includegraphics[width=0.5\textwidth]{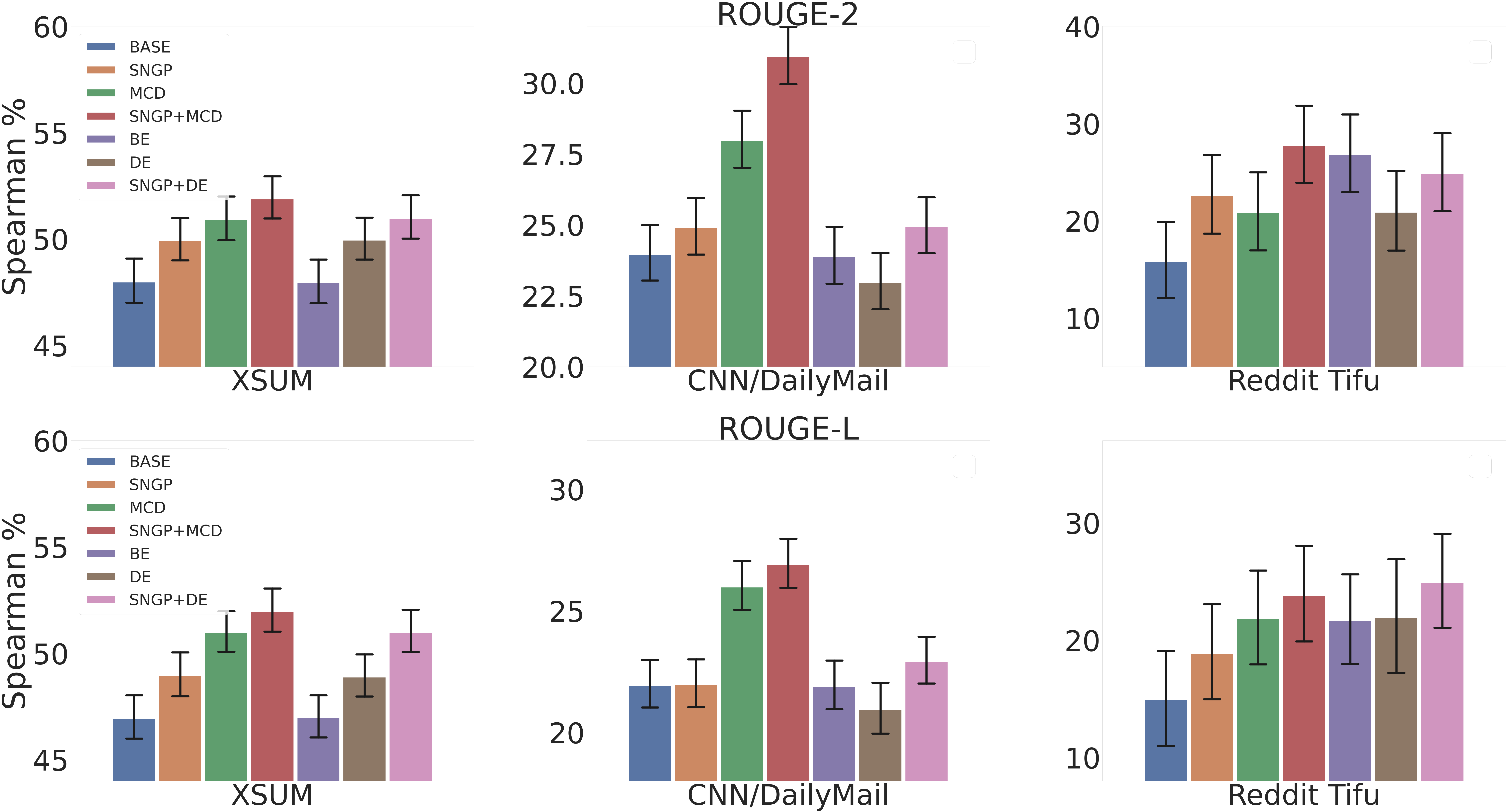}
    }
    \caption{Spearman's rank correlation between the length-normalized log-probabilities and the ROUGE-2 and ROUGE-L.}
    \label{fig:spearman_app}
\end{figure*}

Spearman's rank correlation for the rest of the metrics can be found on Figure~\ref{fig:spearman_app}.
\label{sec:spear_app}

\subsection{Abstention plots}
\begin{figure*}[ht]
    \centering
    \includegraphics[width=0.9\textwidth]{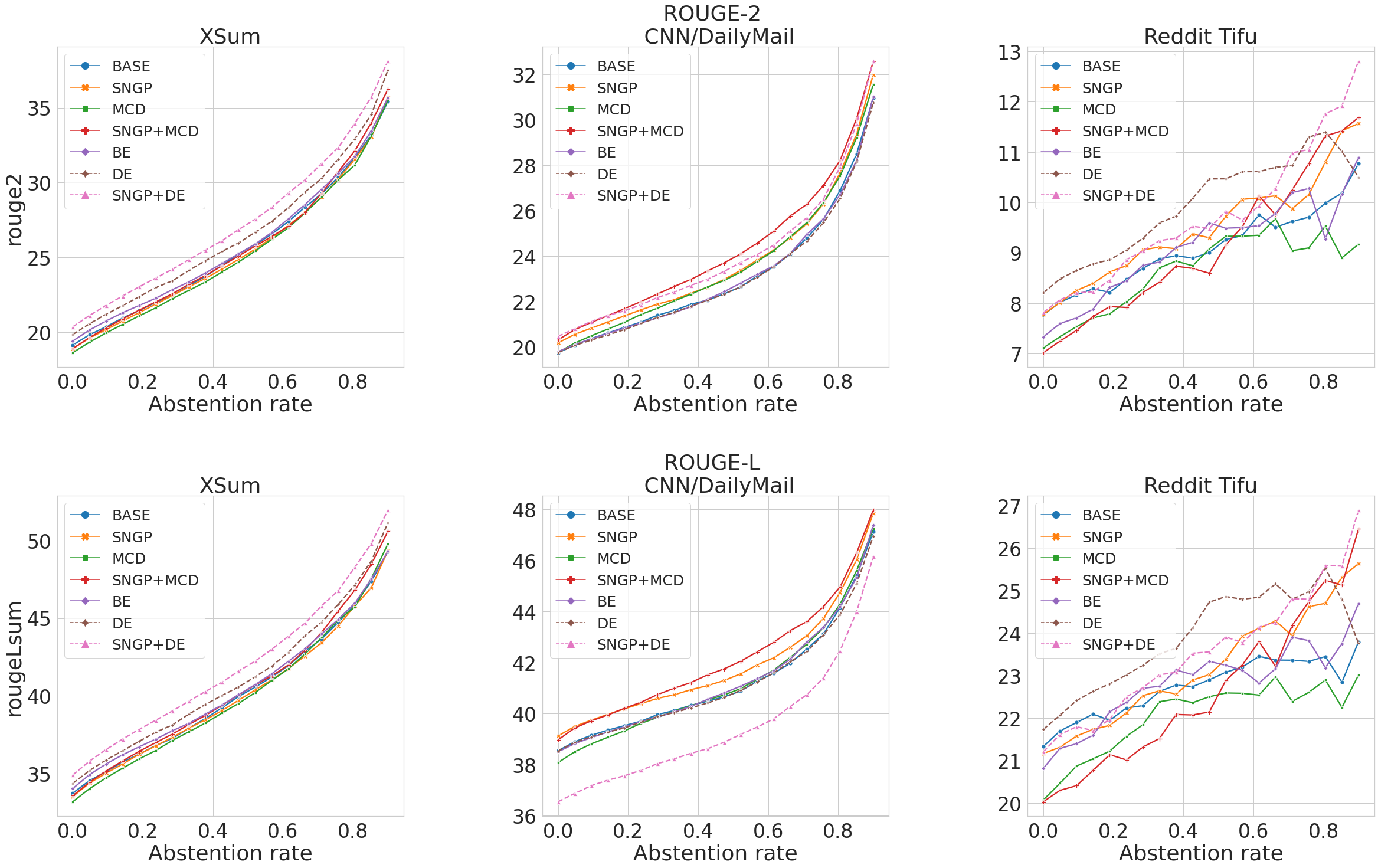}
    \caption{ROUGE-2 and ROUGE-L abstention plots.}
    \label{fig:abstention_app}
\end{figure*}

We demonstrate the abstention plots for the rest of the metrics on Figure~\ref{fig:abstention_app}.
\label{sec:absten_full}

\subsection{Experimental details}
We run all the experiments on the T5 base model (220 million parameters)  using open-sourced T5X framework\footnote{https://github.com/google-research/t5x}.  We used TPU v3 chips. 
Reported metric results are collected from a single evaluation run, unless error bars are provided or stated otherwise.
To select each model checkpoint we ran a hyperparameter sweep to find the best set of parameters. Parameters we sweeped over were: checkpoint step, leaning rate, SNGP mean field factor, number of checkpoints for the ensembles and number of training steps.

In all the experiments we use beam search as a decoding method and we use beam\_size$=3$. For the MCD we used dropout\_rate $= 0.1$ everywhere. Covarience matrix momentum in SNGP was set to $0.999$. For the XSum the best mean field factor $ 10^{-4}$ and for CNN and Reddit it was $10^{-6}$.
\newpage
\subsection{Qualitative results}
\label{sec:quali}
\begin{figure*}[ht]
    \centering
    \includegraphics[width=0.6\textwidth]{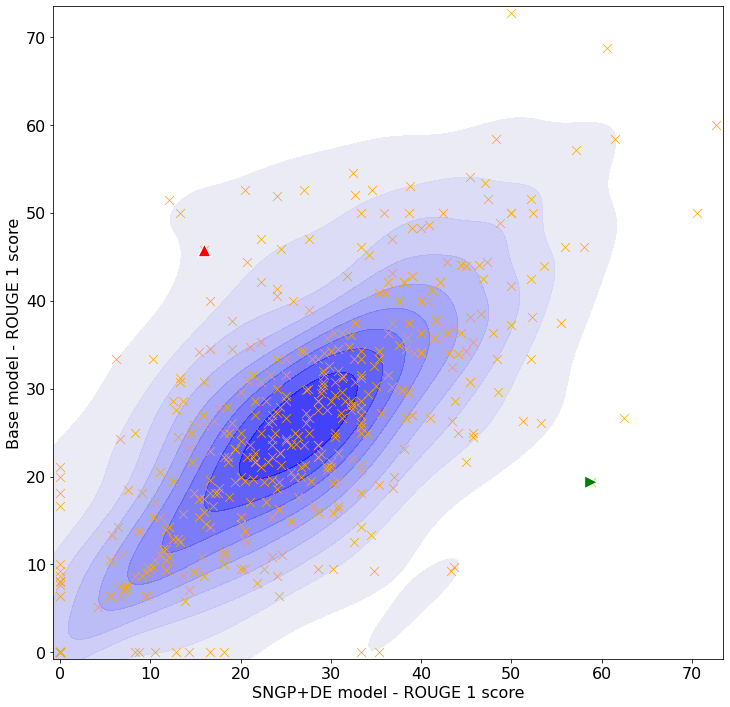}
    \caption{Scatter plot for ROUGE 1 scores of SNGP+DE and Base models on RedditTifu task. Detailed contents of $\textcolor{teal}\blacktriangleright$ and $\textcolor{red}\blacktriangle$ symbols can be founded in Tables \ref{tab:example_red} and \ref{tab:example_teal} respectively.}
    \label{fig:scatter_main}
\end{figure*}

\begin{table}[h]
\centering
\resizebox{0.8\columnwidth}{!}{%
\begin{tabular}{p{\linewidth}}
\toprule
\textbf{Input:} summarize: my family was out on our porch, having some drinks and horderves before giving dad his father's day gift. at some point my mom notices one of our neighbors walking down the street.

she points out that hes wearing these funny shoes and i turn around to see him shuffling his feet while wearing these giant slippers that look like part of a goofy costume. as he's crossing the street to check on his daughter playing at her friends house my dad yells "love those shoes!" and the guy waves and responds "thanks, they're from her," while pointing to his daughter.

him and his daughter exchange a few words and he turns around to walk back home. you have to imagine this guy moving his body side to side while walking, dragging his one foot and swinging the other around, making it look like the slippers were heavy (since they were quite large).

as he passes in front of our house again, i decide to poke a little fun at the way he walked in the slippers and yelled with a chuckle, "hey y'know you walk a little funny in those!" to which he replied, "heh, i dont have much of a choice," and continues walking with his head down. in my brain, he was saying "there's no other way to walk in these damn things," so i laughed audibly so he could hear i appreciated the response and attitude about the silly present.

once he's inside, my mom turns to me and says, "yknow he used to weigh like 300 pounds, he's lost over 120. he also has cerebral palsy so he's always kinda walked with a limp."
\\\toprule
\textbf{Target:} thought a guy's shoes were the reason he walked funny, turns out he has cerebral palsy.
\\\toprule
\textbf{Base model:} i made fun of a guy walking in giant slippers on father's day.
\\\hline
\textbf{SNGP+DE model:} i laughed at the way a guy walked in his slippers, turns out he has cerebral palsy.
\\ \bottomrule
\caption{An example where SNGP+DE model gives better ROUGE1 score than Base model. This is annotated by $\textcolor{teal}\blacktriangleright$ symbol in Figure \ref{fig:scatter_main}.}
\label{tab:example_red}
\end{tabular}
}
\end{table}

\begin{table}[h]
\centering
\resizebox{0.8\columnwidth}{!}{%
\begin{tabular}{p{\linewidth}}
\toprule
\textbf{Input:} summarize: i[f] have small hairs on my lip and waxing doesn't seem to work because of how tiny and thin they are. shaving doesn't help very much since it causes farther irritation. when i was feeling extra self concious after trimming the hairs on valentines day, my fiancé brought up trying the cream hair removal.

i went on amazon and bought the veet hair removal cream. last night, i got it in the mail and read the precautions. i saw not to use it on the face but like an idiot, i thought to ignore it (as does every story about the hair removal cream). it totally did the trick and my lip is hairless. i felt a bit of burning and irritation on the lip after but it went away after using a bit of bio oil. didn't think about it all night.

this morning on the other hand i woke up with more pain that you would feel after a burn. there was a small patch of skin breakdown and irritation to the left of my lip and a bit of redness on my upper lip but nothing more. i covered it up with makeup and it seemed to have done the trick.

fast forward about 8hrs. i now have small pin sized scabs all across my upper lip and pain. i look like a 15 year old boy who doesn't know how to shave or someone with uncontrollable herpes cold sores. ontop of that i got a venus razors ad while i write this on my smart phone to rub it in some more that i should've used a razor. the next week will be ugly.
\\ \toprule 
\textbf{Target}: used hair remover cream on face, now have a chemical burn on the upper lip
\\ \toprule 
\textbf{Base model:} i used veet hair removal cream on my lip and now i have small scabs all over my upper lip.
\\\hline
\textbf{SNGP+DE model:} veet hair removal cream made me look like an idiot.\\
\bottomrule

\caption{An example where SNGP+DE model gives worse ROUGE1 score than Base model. This is annotated by $\textcolor{red}\blacktriangle$ symbol in Figure \ref{fig:scatter_main}.}
\label{tab:example_teal}
\end{tabular}
}
\end{table}




\end{document}